\documentclass[letterpaper,journal]{IEEEtran}
\usepackage{cite}
\usepackage{amsmath,amssymb,amsfonts}
\usepackage{algorithmic}
\usepackage{graphicx}
\usepackage{textcomp}
\usepackage{subcaption}

\IEEEoverridecommandlockouts

\usepackage{multirow}
\usepackage{gensymb}
\usepackage{rotating}
\usepackage{color,soul}
\usepackage{makecell}

\usepackage{booktabs}
\usepackage{bm}

\usepackage[colorlinks=true,linkcolor=black,citecolor=black,urlcolor=black]{hyperref}
\usepackage[capitalize,nameinlink]{cleveref}

\Crefname{section}{Sec.}{Secs.}
\Crefname{figure}{Fig.}{Figs.}
\Crefname{table}{Table}{Tables}
\crefname{equation}{}{}
\Crefname{equation}{}{}
\def\BibTeX{{\rm B\kern-.05em{\sc i\kern-.025em b}\kern-.08em
    T\kern-.1667em\lower.7ex\hbox{E}\kern-.125emX}}

\begin{document}

\title{ReRadar: Robust Radar Global Localization via Rotation-Equivariant Descriptor Learning}

\author{Duc Manh Nguyen$^{1}$*,
Truong Giang Dao$^{1}$*,
Gia Nghiem Luong$^{1}$,
Viet Trung Hoang$^{2}$,
Anh Quang Nguyen$^{1}$
\thanks{$^{1}$ Duc Manh Nguyen, Truong Giang Dao, Gia Nghiem Luong, and Anh Quang Nguyen are with the School of Electrical and Electronic Engineering, Hanoi University of Science and Technology, Vietnam.}
\thanks{$^{2}$ Viet Trung Hoang is with the School of Information and Communications Technology, Hanoi University of Science and Technology, Vietnam.}
\thanks{* Equal contribution.}
\thanks{Corresponding author: quang.nguyenanh@hust.edu.vn.}
}%

\maketitle

\begin{abstract}
Global localization with scanning millimeter-wave radar remains challenging because place-recognition descriptors often discard spatial structure needed for accurate pose retrieval. We present ReRadar, a radar global localization pipeline that extracts rotation-equivariant intermediate features using steerable convolutional neural networks, forms rotation-invariant descriptors through group pooling and NetVLAD aggregation, and combines descriptor retrieval with landmark-based matching to estimate the robot's three-degree-of-freedom (3-DoF) pose.
Across fixed database--query evaluations, ReRadar with target-dataset adaptation achieves 99.37\% Recall@1 on OORD Bellmouth, 91.44\% Recall@1 with 80.99\% $F_1^{\max}$ on Mulran DCC01, and 99.38\% Recall@1 on falling-snow Boreas sequence. Without target-dataset data, the cross-dataset model reaches 98.07\% Recall@1 on OORD, performing comparably to the evaluated state-of-the-art methods.
\end{abstract}




\section{Introduction}
\label{sec:introduction}
Place recognition and pose estimation are fundamental problems in autonomous exploration and long-term operation, especially in Simultaneous Localization and Mapping (SLAM), where they help mitigate accumulated drift through reliable loop-closure constraints~\cite{lim2023orora, zhang20234dradarslam, hong2020radarslam, wang2022maroam}. While extensive research has focused on cameras and LiDAR~\cite{prsurvey}, automotive radar is particularly attractive for real-world deployment because it provides long-range sensing and robust perception under adverse weather, poor lighting, and other visually degraded conditions~\cite{patole2017automotive,hong2020radarslam}.

Recent years have seen substantial progress in radar-based place recognition. Existing methods include handcrafted descriptors~\cite{jang2023raplace, kim2024referee, gadd2024openradvlad} and deep learning approaches. The latter include formulating place recognition as a classification problem~\cite{agarwal2025bayesian}, using NetVLAD~\cite{arandjelovic2016netvlad} with pretrained backbones~\cite{suaftescu2020kidnapped}, and adopting contrastive learning strategies~\cite{gadd2024oord, gadd2021contrastive}.

Although these methods achieve strong place-recognition performance, most are primarily designed to produce global descriptors and do not explicitly preserve spatial information from radar scans. As a result, many existing radar place-recognition methods are not directly applicable to pose retrieval or global localization tasks that require accurate spatial correspondences.

To address this gap, we propose ReRadar, a radar global-localization pipeline that learns translation- and rotation-equivariant features using steerable convolutional neural networks, aggregates them with NetVLAD~\cite{arandjelovic2016netvlad} for robust place retrieval, and performs landmark-based matching for final 3-DoF pose estimation.

In summary, our main contributions are as follows:
\begin{itemize}
    \item We design \textit{ReRadar}, a global localization framework that combines rotation-equivariant descriptor learning and landmark-based geometric matching to recover 3-DoF pose from radar scans.
    \item We introduce a multi-stage training strategy with the InfoNCE objective~\cite{oord2018representation} for descriptor learning, and we detail how this training pipeline improves robustness under cross-environment localization.
    \item We evaluate the method on { OORD, Mulran{, and Boreas}~{\cite{gadd2024oord,kim2020mulran,burnett2023boreas} using fixed database--query configurations under both in-dataset and cross-dataset training settings}, showing {strong place-recognition }and {pose-estimation performance}}.
    \item We validate embedded deployment on NVIDIA Jetson Orin Nano with an end-to-end runtime of 294.98\,ms per frame, showing near real-time operation for practical autonomous systems.
\end{itemize}

\section{Related Work}

\subsection{Radar-based Place Recognition}
Radar-based place recognition methods can be broadly categorized according to their sensor inputs into single-chip radar and scanning radar approaches~\cite{kim2024referee}.

For single-chip radar, AutoPlace~\cite{cai2022autoplace} combines measurements from five automotive radars and employs dynamic-point removal, spatiotemporal feature embedding, and candidate refinement for place recognition. mmPlace~\cite{meng2024mmplacerobustplacerecognition} transforms intermediate-frequency radar signals into range--azimuth heatmaps and extracts spatial features from concatenated heatmaps acquired using a rotating radar platform. SPR~\cite{SPR} provides a lightweight place-recognition framework that operates on a single-chip radar scan. These methods address the limitations of single-chip radar measurements through multi-sensor integration, enhanced radar representations, and lightweight descriptor learning.

In contrast, ReRadar operates on polar intensity images acquired by a 360$^\circ$ scanning radar and is not directly applicable to the sparse automotive-radar point-cloud inputs provided by datasets such as nuScenes~\cite{caesar2020nuscenes}. For scanning-radar place recognition, RaPlace~\cite{jang2023raplace} uses the Radon transform to capture global linear structures and achieve rotation robustness without learning. RingKey, as defined in the OORD benchmark~\cite{gadd2024oord}, uses the ring-key component of Scan Context~\cite{kim2018scan}: observations at each fixed radial distance are averaged to form a compact rotation-invariant descriptor, without applying Scan Context's orientation-refinement stage.

RaPlace requires a polar-to-Cartesian conversion before descriptor extraction. Open-RadVLAD~\cite{gadd2024openradvlad} (hereafter referred to as RadVLAD) instead operates directly on polar radar data. It applies a 1D Fourier transform along radar radial beams and aggregates the resulting frequency responses using VLAD (Vector of Locally Aggregated Descriptors)~\cite{VLAD}. These approaches illustrate different choices of representation and aggregation for constructing global radar descriptors.

Learning-based approaches provide an alternative to handcrafted descriptors. Gadd \textit{et al.}~\cite{gadd2021contrastive} employ contrastive learning to obtain discriminative radar representations. The neural baseline in OORD~\cite{gadd2024oord} combines a ResNet18 backbone~\cite{he2016deep} with NetVLAD aggregation~\cite{arandjelovic2016netvlad}. This architecture aggregates local deep features into a fixed-length global descriptor for place retrieval; its initialization and adaptation settings in our evaluation are specified in Section~\ref{sec:setting}.

Other learning-based methods include Kidnapped~\cite{suaftescu2020kidnapped}, which uses NetVLAD with a VGG16 backbone~\cite{simonyan2015deepconvolutionalnetworkslargescale}, and the approach of Agarwal \textit{et al.}~\cite{agarwal2025bayesian}, which formulates place recognition as a classification problem. These methods primarily target global place recognition rather than explicit local correspondence estimation. Consequently, retrieving a candidate place does not by itself recover the relative pose between the query and the corresponding map observation, motivating the integration of retrieval with geometric matching.

\subsection{Radar Scan Matching}

Radar scan matching estimates relative motion by registering radar observations using geometric or feature correspondences. Cen \textit{et al.}~\cite{cen2018precise} introduce a radar-specific landmark detector and matching pipeline for ego-motion estimation, and subsequently improve correspondence estimation through graph matching~\cite{cen2019radar}. These methods provide radar-specific mechanisms for detecting landmarks and establishing correspondences between observations.

CFEAR Radarodometry~\cite{adolfsson2021cfear} extracts a sparse set of oriented surface points by conservatively filtering radar returns and registers scans against a history of keyframes using a robust point-to-line objective. ORORA~\cite{lim2023orora} addresses erroneous radar correspondences by decoupling rotation and translation estimation while accounting for the anisotropic uncertainty of radar measurements.

These methods primarily focus on local alignment between temporally or spatially nearby observations. Their registration mechanisms can complement place-recognition systems by estimating the relative pose after a candidate map observation has been retrieved. ReRadar adopts this combined strategy: global descriptor retrieval identifies a candidate radar scan, followed by radar-specific landmark matching and RANSAC-based 3-DoF pose estimation.

\subsection{Radar-based Place Recognition with Pose Estimation}

Global localization requires more than identifying a previously observed place: it also requires estimating the pose of the current observation relative to the map. Existing approaches differ in their map modality and in how they combine place recognition with metric pose estimation.

In radar-to-LiDAR localization, RaLL~\cite{Yin2020RaLLER} localizes radar observations directly in a LiDAR map using a differentiable measurement model combined with a Kalman filter to exploit temporal continuity during pose estimation. RaLF~\cite{Nayak2023RaLFFG} instead formulates metric localization as a flow estimation problem, predicting pixel-level flow between radar and LiDAR image representations to infer the 3-DoF pose transformation. RoLM~\cite{ma2023rolm} employs a Scan Projection Descriptor (SPD) to represent radar scans and LiDAR maps in a unified density-descriptor space. These approaches use cross-modal representations to connect radar observations to a LiDAR reference map.

Radar-only localization avoids dependence on a pre-built LiDAR map. RadarLoc~\cite{Wang2021RadarLocLT} jointly models global and relative pose to capture spatial structure and incorporates a self-attention mechanism to mitigate the effects of noise and dynamic objects in urban radar environments. ReFeree~\cite{kim2024referee} exploits free-space information to construct descriptors for place recognition and initial-heading estimation, rather than directly estimating the full translational and rotational pose.

ReRadar also addresses radar-only localization, combining global retrieval with local landmark correspondences to recover the full 3-DoF pose. The retrieval stage selects a candidate map observation, while geometric matching estimates the relative transformation between that observation and the query. This separation allows place-recognition performance and geometric localization reliability to be evaluated explicitly.

\section{Method}
\label{sec:methodology_intro}
\begin{figure*}[t!]
    \centering
    \includegraphics[width=0.96\linewidth]{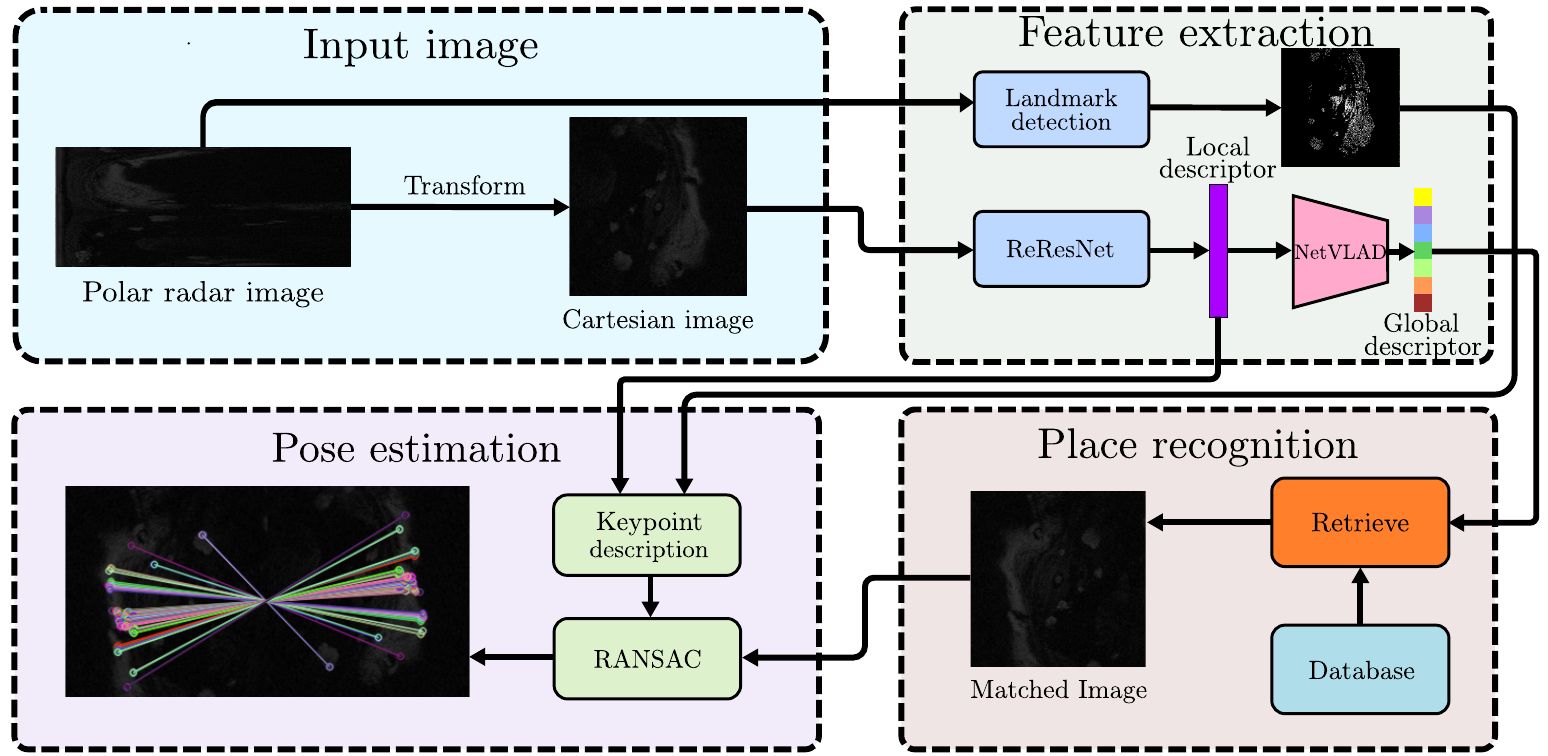}
    \caption{Overview of the proposed pipeline. A polar radar image is first transformed into a Cartesian image and then fed into ReResNet~\cite{han2021redet} to extract local descriptors. In parallel, the polar radar image is processed by a landmark detection module to detect keypoints. NetVLAD~\cite{arandjelovic2016netvlad} {is} then used to aggregate features into a global descriptor. When the global descriptors of a query image and a database image are matched, the corresponding precomputed keypoint descriptors are used to estimate their relative pose via RANSAC~\cite{fischler1981random}.}
    \label{fig:pipeline}
\end{figure*}


\subsection{Global Localization Pipeline}
\label{pipeline}

Given a query radar scan $I_q$, the global localization task is decomposed into two main steps: (1) place recognition to retrieve the most similar database entry, and (2) relative pose estimation between the query and the retrieved entry.

We first extract the global descriptor $\mathbf{d}_q$ of the query scan (detailed in Section~\ref{sec:feature_extraction}). During offline database construction, the global descriptor $\mathbf{d}_i$ of every keyframe radar image is precomputed using the same descriptor extractor. The best-matching database index $\hat{m}$ is then retrieved by nearest-neighbor search as follows:
\begin{equation}
\hat{m} = \arg\min_{m \in \{1, \dots, {N_{\mathrm{db}}}\}} \|\mathbf{d}_q - \mathbf{d}_m\|_2.
\label{eq:retrieval}
\end{equation}
{where $N_{\mathrm{db}}$ is the number of database entries.}

The relative 2D rigid-body transformation between the query scan $I_q$ and the retrieved database scan $I_{\hat{m}}$ is subsequently estimated by the pose estimation module as follows:

\begin{equation}
\hat{\mathbf{T}}_{q \to \hat{m}} = \mathrm{PoseEst}(I_q, I_{\hat{m}}),
\label{eq:poseest}
\end{equation}
where $\hat{\mathbf{T}}_{q \to \hat{m}} = [\hat{x}, \hat{y}, \hat{\theta}]^\top$ denotes the estimated relative translation and heading angle. The function $\mathrm{PoseEst}(\cdot)$ performs keypoint detection, local descriptor assignment, correspondence matching, and robust RANSAC-based transformation estimation (detailed in Section~\ref{sec:pose}).

\subsubsection{Rotation-Equivariant and Invariant Descriptors}
\label{sec:feature_extraction}

For robust place recognition under arbitrary heading changes, {we learn intermediate feature maps that are equivariant to translation and rotation, and subsequently transform them into a rotation-invariant global descriptor}. Standard convolutions are naturally translation-equivariant in Cartesian coordinates~\cite{luo2025bevplace++}, whereas polar representations effectively convert this property into rotation equivariance; however, due to non-uniform sampling, they often degrade translation consistency~\cite{esteves2017polar, serio2025polarperspectivesevaluating2d}. Therefore, we operate on Cartesian radar images, which preserve the translation-equivariant behavior of standard CNNs and are more compatible with conventional convolutional architectures~\cite{luo2025bevplace++}, then explicitly learn rotation-equivariant representations.

We adopt a ResNet-style backbone~\cite{he2016deep} and replace standard convolutions with steerable convolutions~\cite{weiler2019general}, yielding a ReResNet-based rotation-equivariant encoder~\cite{han2021redet}. Given a Cartesian radar input $I \in \mathbb{R}^{H \times W}$, the encoder produces rotation-equivariant feature maps as follows:
\begin{equation}
F \in \mathbb{R}^{H' \times W' \times C \times |G|},
\end{equation}
where $H'$ and $W'$ denote the spatial resolution, $C$ is the number of base feature channels, and $|G|$ is the number of discrete rotations in the group.
  {Group }pooling {removes }the {explicit orientation index, producing orientation-invariant channel vectors while retaining the spatially equivariant arrangement of the feature map:}
\begin{equation}
F' \in \mathbb{R}^{H' \times W' \times C}.
\end{equation}

The map $F'$ is flattened into $M = H'W'$ local descriptors $\{\mathbf{f}_n \in \mathbb{R}^C\}_{n=1}^M$. These descriptors are aggregated using NetVLAD~\cite{arandjelovic2016netvlad} with $K$ learnable cluster centers $\{\mathbf{c}_k \in \mathbb{R}^C\}_{k=1}^K$. The $k$-th VLAD vector is computed as follows:
\begin{equation}
\mathbf{v}_k = \sum_{n=1}^M a_{nk} (\mathbf{f}_n - \mathbf{c}_k),
\end{equation}
where $a_{nk}$ denotes the soft-assignment weight. The final global descriptor is defined as follows:
\begin{equation}
\mathbf{d} = [\mathbf{v}_1, \dots, \mathbf{v}_K] \in \mathbb{R}^{CK},
\end{equation}
{Under an input rotation, the spatial positions in $F'$ rotate accordingly, whereas their channel representation is invariant to the pooled orientation index. Because NetVLAD aggregates the local descriptors over spatial positions without preserving their ordering, the resulting global descriptor $\mathbf{d}$ is rotation-invariant. Thus, equivariance refers to the intermediate spatial feature maps, while invariance refers to the final descriptor used for retrieval.}

\subsubsection{Pose Estimation}
\label{sec:pose}

\begin{figure}[t!]
    \centering
    \begin{minipage}{0.32\linewidth}
        \centering
        \includegraphics[width=\linewidth]{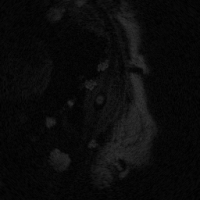}
        \footnotesize (a) Radar scan
    \end{minipage}
    \hfill
    \begin{minipage}{0.32\linewidth}
        \centering
        \includegraphics[width=\linewidth]{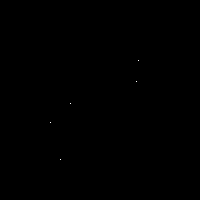}
        \footnotesize (b) SIFT~\cite{lowe2004distinctive}
    \end{minipage}
    \hfill
    \begin{minipage}{0.32\linewidth}
        \centering
        \includegraphics[width=\linewidth]{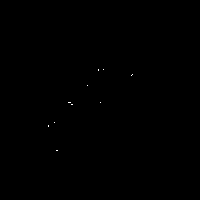}
        \footnotesize (c) ORB~\cite{rublee2011orb}
    \end{minipage}
    \vspace{0.5em}
    \begin{minipage}{0.32\linewidth}
        \centering
        \includegraphics[width=\linewidth]{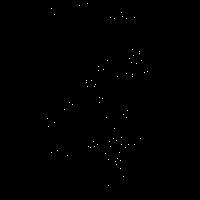}
        \footnotesize (d) FAST~\cite{rosten2006machine}
    \end{minipage}
    \hfill
    \begin{minipage}{0.32\linewidth}
        \centering
        \includegraphics[width=\linewidth]{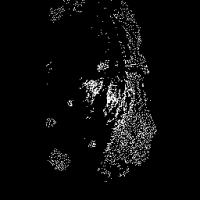}
        \footnotesize (e) Cen~\cite{cen2018precise}
    \end{minipage}
    \hfill
    \begin{minipage}{0.32\linewidth}
        \centering
        \includegraphics[width=\linewidth]{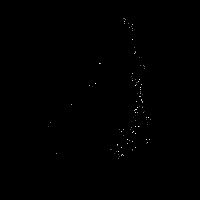}
        \footnotesize (f) Cen~\cite{cen2019radar}
    \end{minipage}
    \caption{Qualitative comparison of keypoint detection methods on OORD~\cite{gadd2024oord} radar scans. Traditional detectors (SIFT~\cite{lowe2004distinctive}, ORB~\cite{rublee2011orb}, and FAST~\cite{rosten2006machine}) yield sparse and unstable keypoints under speckle noise and low-texture conditions, whereas radar-specific landmark detectors (Cen \textit{et al.}~\cite{cen2018precise, cen2019radar}) produce denser and more repeatable keypoints, demonstrating their suitability for robust pose estimation.}
    \label{fig:keypoint_comparison}
\end{figure}

{Reliable relative-pose estimation requires stable keypoints and local descriptors. As illustrated in Fig.~\ref{fig:keypoint_comparison}, FAST~\cite{rosten2006machine}, SIFT~\cite{lowe2004distinctive}, and ORB~\cite{rublee2011orb}, which target textured camera images, perform poorly on sparse, noisy radar scans, whereas the radar-specific detectors of Cen \textit{et al.}~\cite{cen2018precise,cen2019radar} are more repeatable under speckle noise and sparsity. Among the two radar-specific detectors, we adopt the method of Cen \textit{et al.}~\cite{cen2018precise} because it provides more usable keypoints than the method in~\cite{cen2019radar}, as illustrated in Fig.~\ref{fig:keypoint_comparison}.}

{Specifically, each detected keypoint is mapped to its corresponding discrete location on the spatially equivariant feature map $F'$, whose channel vectors are orientation-invariant (Section~\ref{sec:feature_extraction}), and its local descriptor is obtained by direct indexing.} Tentative correspondences are then established by nearest-neighbor descriptor matching in descriptor space.

Let
\(\mathcal{M} = \{(\mathbf{u}_q^j, \mathbf{u}_{\hat{m}}^j)\}_{j=1}^J\)
denote the set of matched keypoints between the query image $I_q$ and the retrieved database image $I_{\hat{m}}$, where
\(\mathbf{u}_q^j=(u_q^j,v_q^j)^\top\) and
\(\mathbf{u}_{\hat{m}}^j=(u_{\hat{m}}^j,v_{\hat{m}}^j)^\top\)
{are 2D coordinates in the metric Cartesian BEV frame. Before RANSAC, the detected pixel coordinates are converted to meters using the radar-image resolution of $0.64\,\mathrm{m/pixel}$.}

We estimate a planar rigid transform
\(\mathcal{T}_{[\mathbf{t},\theta]}\)
with translation
\(\mathbf{t}=[\hat{x},\hat{y}]^\top\)
and rotation
\(\theta\in[0,360)^\circ\) as follows:
\begin{equation}
\mathcal{T}_{[\mathbf{t},\theta]}(\mathbf{u}) = {\mathbf{R}(\theta)}\,\mathbf{u} + \mathbf{t},
\end{equation}
where {$\mathbf{u}\in\mathbb{R}^2$ is a 2D metric BEV coordinate}, $\mathbf{t}$ is expressed in meters, and
${\mathbf{R}(\theta)}$ is the corresponding $2\times2$ rotation matrix.
The optimal transformation is found using RANSAC~\cite{fischler1981random} by maximizing the number of inliers as follows:
\begin{equation}
{
\hat{\mathcal{T}}_{q\rightarrow\hat{m}}
=\arg\max_{\mathcal{T}_{q\rightarrow\hat{m}}}
\sum_{j=1}^{J}
\mathbf{1}\left[
\left\|
\mathbf{u}_{\hat{m}}^{j}
-\mathcal{T}_{q\rightarrow\hat{m}}(\mathbf{u}_q^{j})
\right\|_2
\leq\epsilon
\right],}
\end{equation}
where $\epsilon > 0$ is the inlier threshold. Finally, the relative pose estimate $[\hat{x}, \hat{y}, \hat{\theta}]^\top$ is extracted from ${\hat{\mathcal{T}}_{q\rightarrow\hat{m}}}$.

\subsection{Network Training}
\label{training}

\subsubsection{Training Strategy}

\begin{figure*}[t!]
    \centering
    \includegraphics[width=0.96\textwidth]{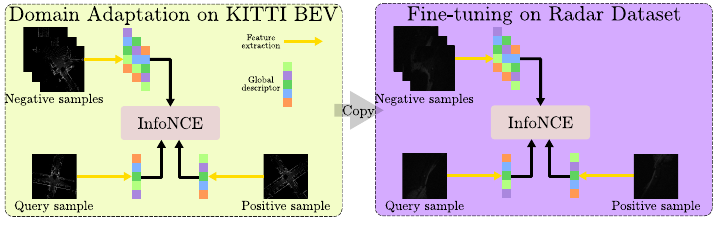}
    \caption{Overview of the proposed two-stage training pipeline. Stage 1 performs intermediate pretraining on KITTI sequence 00~\cite{geiger2013vision} from ImageNet-pretrained initialization, and Stage 2 fine-tunes on the target radar dataset using InfoNCE loss~\cite{oord2018representation}.}
    \label{fig:training}
\end{figure*}

{Training from scratch is computationally expensive and requires substantial data. To improve convergence and data efficiency, we adopt the two-stage strategy illustrated in Fig.~\ref{fig:training}. The ReResNet backbone~\cite{han2021redet} is initialized with ImageNet-pretrained weights~\cite{russakovsky2015imagenet}; Cartesian inputs are processed by ReResNet, and the resulting features are aggregated into global descriptors using NetVLAD~\cite{arandjelovic2016netvlad}. In the first stage, the model is adapted on BEV images from KITTI sequence 00~\cite{geiger2013vision} using the InfoNCE loss~\cite{oord2018representation} with sampled positive and negative descriptor pairs. The KITTI-adapted model is then fine-tuned on radar data using the same objective.}

\subsubsection{Loss Function}
To train ReResNet for place recognition, we optimize the InfoNCE loss ~ \cite{oord2018representation} on global descriptors. For each query descriptor
$\mathbf{d}_q$, with a positive descriptor $\mathbf{d}^+$ and $N_{\mathrm{neg}}$ negative
descriptors $\{\mathbf{d}_j^-\}_{j=1}^{N_{\mathrm{neg}}}$, the loss is defined as follows:
\begin{equation}
\mathcal{L}
=
-\log
\frac{%
\exp\!\left( \frac{\mathbf{d}_q^\top \mathbf{d}^{+}}{\tau} \right)
}{%
\exp\!\left( \frac{\mathbf{d}_q^\top \mathbf{d}^{+}}{\tau} \right)
+
\sum_{j=1}^{N_{\mathrm{neg}}}
\exp\!\left( \frac{\mathbf{d}_q^\top \mathbf{d}_j^{-}}{\tau} \right)
}
\end{equation}

where all descriptors are $\ell_2$-normalized, so inner products correspond to cosine similarity. The positive sample is selected within a predefined spatial radius of the query, while negatives are sampled outside that radius. The temperature $\tau$ controls similarity concentration, and the objective encourages higher similarity for positive pairs than for negatives.

\section{Experiments}
\label{sec:experiment}

\subsection{Dataset}
\label{sec:dataset}
{To evaluate the proposed pipeline under diverse operating conditions, we used three public datasets: OORD~\cite{gadd2024oord} for off-road environments, Mulran~\cite{kim2020mulran} for urban scenes, and Boreas~\cite{burnett2023boreas} for long-term seasonal and weather variations.}
\subsubsection{OORD}
\label{sec:dataset_oord}
The OORD dataset consists of radar data
collected in four off-road environments in rural regions of
England, where each location is traversed multiple times
along the same routes at different times. This characteristic
makes OORD well suited for multi-session place recognition
evaluation. In this paper, we used the trajectory pairs defined
by the dataset benchmark: {Bellmouth 2 vs Bellmouth 1, Hydro 1 vs Hydro 2, Hydro 1 vs Hydro 3, Maree 1 vs Maree 2, and Two Lochs 2 vs Two Lochs 1.}

 \subsubsection{Mulran}
\label{sec:dataset_mulran}
For Mulran, we used a single-session evaluation protocol and split each sequence into database and query subsets. {The DCC01 sequence was split at scan 765, while KAIST03 was split at scan 1420; scans before each split were used to construct the database, and the subsequent scans formed the query set.} Mulran contains long radar sequences with multiple loops and diverse dynamic objects, providing a challenging setting for evaluating place recognition performance under repetitive structures and dynamic environments.

 \subsubsection{\texorpdfstring{Boreas}{Boreas}}
\label{sec:dataset_boreas}
{Boreas contains 44 driving sequences spanning more than 350\,km, primarily repeated traversals of the Glen Shields route near Toronto over one year, together with routes in unseen environments. Its 360$^\circ$ scanning-radar data cover substantial seasonal and weather variations, including rain and snow, supporting long-term place-recognition evaluation.}
{We used the sunny sequence \texttt{2021-03-02-13-38} as the database and the snow-covered \texttt{2020-11-26-13-58} and falling-snow \texttt{2021-01-26-11-22} sequences as queries.}

\subsection{\texorpdfstring{Radar-Domain Training Settings}{Radar-Domain Training Settings}}
\label{sec:generalization_protocols}

{All experiments used the fixed database--query configurations in Section~\ref{sec:dataset}. Our (1) was trained only on KITTI~\cite{geiger2013vision}. Our (2) was adapted from Our (1) using only non-target radar data: one loop from each of the Mulran sequences \texttt{KAIST02}, \texttt{DCC02}, \texttt{Riverside01}, and \texttt{Sejong03} for the OORD and Boreas models, and Boreas sequences \texttt{2021-05-06-13-19}, \texttt{2020-12-01-13-26}, and \texttt{2021-01-19-15-08} for the Mulran model. Our (3) was instead adapted using target-dataset data: \texttt{Two Lochs 2} for OORD, \texttt{KAIST02} for Mulran, and \texttt{boreas-2020-12-01-13-26} for Boreas. None was used as a query sequence; however, \texttt{Two Lochs 2} also served as the database for its evaluation pair, making that setting map-adapted rather than scene-disjoint.}

 \subsection{Implementation Details}
\label{sec:setting}
For the baseline methods RaPlace~\cite{jang2023raplace}, RadVLAD~\cite{gadd2024openradvlad}, RingKey~\cite{kim2018scan}, ReFeree~\cite{kim2024referee}, and ResNet18-NetVLAD~\cite{gadd2024oord},
we followed the experimental settings reported in~\cite{gadd2024oord}
to ensure a fair comparison.
{For ResNet18-NetVLAD, the ImageNet-pretrained backbone was not fine-tuned on target radar data; following OORD~\cite{gadd2024oord}, only the NetVLAD centers were initialized by k-means++ on deep features from the reference trajectory.}

For our method, we used ReResNet-50~\cite{han2021redet} up to the second stage, implemented using the rotational-equivariant layers of $E(2)$-CNN~\cite{weiler2019general} with the cyclic group $G_8$, producing feature maps with $C = 512$ channels.
{Following~\cite{gadd2021contrastive}, the input polar radar images were first converted to $200 \times 200$ Cartesian images at a resolution of 0.64\,m/pixel, covering a spatial range of 128\,m.} We then trained using the InfoNCE loss~\cite{oord2018representation} with temperature $\tau = 0.1$, 14 negative samples per query, a batch size of 2, and 30 epochs.

{For the backbone ablation, Our (2)-ResNet and Our (3)-ResNet used the corresponding settings in Section~\ref{sec:generalization_protocols}, replacing ReResNet with a standard ResNet. All models were trained on an NVIDIA RTX 3090 GPU with learning rates of $10^{-4}$ from ImageNet initialization~\cite{russakovsky2015imagenet} and $10^{-5}$ when adapting the KITTI-pretrained model.}

\subsection{Evaluation Metrics}
\label{sec:metrics}
{Following prior radar place-recognition studies~\cite{suaftescu2020kidnapped,jang2023raplace}, we report Recall@1:}
\begin{equation}
\mathrm{Recall@1} = \frac{TP}{GT},
\end{equation}
{A query is counted as a true positive (TP) when its top-ranked retrieval lies within 25\,m of the ground truth on OORD~\cite{gadd2024oord}, or within 20\,m on Mulran and Boreas~\cite{kim2020mulran,burnett2023boreas}; GT is the total number of queries.}

{To summarize the precision--recall trade-off independently of a fixed descriptor-distance threshold, we also report}
\begin{equation}
F_1^{\max} = \max_{\delta} \frac{2\,P(\delta)\,R(\delta)}{P(\delta)+R(\delta)},
\end{equation}
{where $P(\delta)$ and $R(\delta)$ are precision and recall at threshold $\delta$. For geometric accuracy, we report mean translation error $\hat{e_t}$ in meters and rotation error $\hat{e_r}$ in degrees, both computed over successful cases only.}

 { {Following BEVPlace++~\cite{luo2025bevplace++}, we additionally report localization success rate (SR). A localization succeeds when translation and rotation errors are jointly below $(25\,\mathrm{m},10^\circ)$ on OORD or $(20\,\mathrm{m},10^\circ)$ on Mulran and Boreas. Since ReFeree~\cite{kim2024referee} does not estimate translation, its success requires correct place retrieval and rotation error below $10^\circ$:}}
\begin{equation}
{\mathrm{SR}=\frac{N_{\mathrm{succ}}}{GT}.}
\end{equation}
{where $N_{\mathrm{succ}}$ is the number of successful localizations.}
 \section{Results}
\label{sec:results}
 {The following experiments used the same database--query pairs for all three ReRadar settings, allowing the effects of KITTI-only training, cross-dataset radar adaptation, and target-dataset radar adaptation to be compared directly.
}
\subsection{Results on OORD}
\label{sec:oord}


\begin{table*}[t!]
\centering

\caption{{Quantitative comparison on OORD~\cite{gadd2024oord}. The ReRadar settings and evaluation metrics are defined in Sections~\ref{sec:setting} and~\ref{sec:metrics}, respectively. Best values are shown in bold.}}
\hypersetup{linkcolor=black,citecolor=black}
\resizebox{\textwidth}{!}{
\renewcommand{\arraystretch}{1.2}
\begin{tabular}{c|ccccc|ccccc|ccccc|ccccc|ccccc}
 & \multicolumn{5}{c|}{Bellmouth 2-vs-Bellmouth 1}
 & \multicolumn{5}{c|}{Hydro 1-vs-Hydro 2}
 & \multicolumn{5}{c|}{Hydro 1-vs-Hydro 3}
 & \multicolumn{5}{c|}{Maree 1-vs-Maree 2}
 & \multicolumn{5}{c}{Two Lochs 2-vs-Two Lochs 1} \\

Method
& Recall@1 & $F_1^{\max}$ & {SR} & $\hat{e_r}$ & $\hat{e_t}$
& Recall@1 & $F_1^{\max}$ & {SR} & $\hat{e_r}$ & $\hat{e_t}$
& Recall@1 & $F_1^{\max}$ & {SR} & $\hat{e_r}$ & $\hat{e_t}$
& Recall@1 & $F_1^{\max}$ & {SR} & $\hat{e_r}$ & $\hat{e_t}$
& Recall@1 & $F_1^{\max}$ & {SR} & $\hat{e_r}$ & $\hat{e_t}$ \\

\toprule
RaPlace~\cite{jang2023raplace}
& 98.45 & 98.48 & {--} & -- & -- & 92.65 & 94.69 & {--} & -- & -- & 95.34 & 96.96 & {--} & -- & -- & 94.28 & 95.58 & {--} & -- & -- & 85.82 & 87.61 & {--} & -- & -- \\

RadVLAD~\cite{gadd2024openradvlad}
& 98.87 & 99.15 & {--} & -- & -- & 93.51 & 95.79 & {--} & -- & -- & 96.00 & \textbf{97.86} & {--} & -- & -- & 96.88 & 96.75 & {--} & -- & -- & 84.27 & 83.83 & {--} & -- & -- \\

ResNet18-NetVLAD~\cite{gadd2024oord}
& 97.67 & 97.67 & {--} & -- & -- & 93.78 & 96.15 & {--} & -- & -- & 95.70 & 97.15 & {--} & -- & -- & 97.14 & 97.14 & {--} & -- & -- & 5.41 & 5.44 & {--} & -- & -- \\

ReFeree~\cite{kim2024referee}
& 89.29 & 88.79 & {73.03} & {2.52} & -- & 90.13 & 90.13 & {82.10} & {2.41} & -- & 91.55 & 92.37 & {80.59} & {2.6} & -- & 88.69 & 88.68 & {77.76} & {2.95} & -- & 70.75 & 69.75 & {53.61} & {3.83} & -- \\

RingKey~\cite{kim2018scan}
& 94.29 & 94.29 & {--} & -- & -- & 91.15 & 92.66 & {--} & -- & -- & 94.06 & 94.12 & {--} & -- & -- & 91.76 & 91.76 & {--} & -- & -- & 72.96 & 74.59 & {--} & -- & -- \\

\midrule
{\textbf{Our (1)}}
& 96.41 & 96.41 & 85.00 & 2.58 & 1.98
& 92.55 & 94.24 & 88.00 & 2.32 & 1.63
& 96.01 & 96.03 & 86.00 & 2.27 & 1.60
& 92.89 & 93.58 & 83.00 & 2.80 & 2.36
& 87.21 & 83.64 & 65.00 & \textbf{3.74} & 7.05 \\

{\textbf{Our (2)}}
& 96.13 & 96.19 & 83.00 & 2.37 & \textbf{1.63}
& 98.07 & 98.07 & 87.00 & 2.35 & 1.62
& \textbf{97.70} & 97.75 & 86.00 & 2.28 & 1.56
& 96.36 & 96.28 & 84.00 & 2.85 & 2.37
& 84.74 & 80.02 & 65.00 & 3.78 & 7.63 \\

{\textbf{Our (3)}}
& 99.37 & 99.37 & 87.00 & 2.42 & 2.07
& \textbf{98.71} & \textbf{98.71} & 88.00 & 2.31 & 1.87
& 97.24 & 97.81 & 87.00 & 2.37 & 1.78
& 97.90 & 97.76 & \textbf{85.00} & 2.83 & 2.47
& \textbf{96.97} & \textbf{94.41} & \textbf{73.00} & 3.76 & 6.20 \\

{\textbf{Our (2)-ResNet}}
& 84.30 & 84.35 & 70.00 & \textbf{2.34} & 1.69
& 91.32 & 91.38 & 84.00 & 2.39 & \textbf{1.46}
& 91.30 & 91.34 & 83.00 & 2.33 & \textbf{1.47}
& 88.66 & 88.60 & 78.00 & 2.74 & \textbf{2.21}
& 31.69 & 34.35 &  16.00 & 3.85 & 6.51 \\

{\textbf{Our (3)-ResNet}}
& \textbf{99.58} & \textbf{99.58} & \textbf{88.00} & 2.38 & 1.97
& 98.02 & 98.02 & \textbf{89.00} & \textbf{2.27} & 1.67
& 96.42 & 97.73 & \textbf{88.00} & \textbf{2.26} & 1.68
& \textbf{97.99} & \textbf{97.84} & \textbf{85.00} & \textbf{2.69} & 2.29
& 92.46 & 87.24 & 69.00 & 3.80 & \textbf{5.62} \\

\bottomrule
\end{tabular}
}

\label{tab:oord_result}
\end{table*}

{We first examine how radar-domain adaptation transfers across the five OORD trajectory pairs in Table~\ref{tab:oord_result}. Our (1), trained only on KITTI, is already competitive on most pairs, but its performance decreases on the more challenging Two Lochs route. Cross-dataset adaptation in Our (2) is particularly effective on Hydro, increasing Recall@1 to 98.07\% on Hydro (1-vs-2) and 97.70\% on Hydro (1-vs-3), despite using no OORD data for adaptation.}

{Target-dataset adaptation provides the most consistent retrieval performance. Our (3) achieves the highest Recall@1 among the external baselines on all five pairs and the highest $F_1^{\max}$ on four; only on Hydro (1-vs-3) is its $F_1^{\max}$ marginally below RadVLAD. The largest gain occurs on Two Lochs, where Recall@1 and SR increase from 84.74\% and 65\% for Our (2) to 96.97\% and 73\% for Our (3). Across the equivariant ReRadar settings, rotation errors remain within 2.27$^\circ$--3.78$^\circ$, while Two Lochs retains the largest translation error.}

\subsection{  \texorpdfstring{Results on Mulran}{Results on Mulran} }
\label{sec:mulran}

\begin{figure}[t!]
    \centering
    \includegraphics[width=0.48\textwidth]{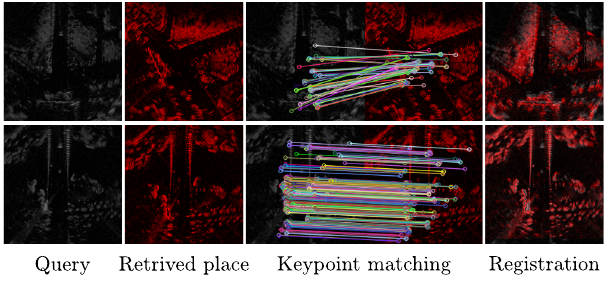}
    \caption{Visualization of registration results produced by the
    {target-dataset-adapted ReRadar model, Our (3),} on the Mulran KAIST03 sequence~\cite{kim2020mulran}.
    The top row shows successful registration under rotation,
    while the bottom row shows successful registration under
    translation.}
    \label{fig:registration}
\end{figure}

\begin{figure}[t!]
    \centering
    \includegraphics[width=0.48\textwidth]{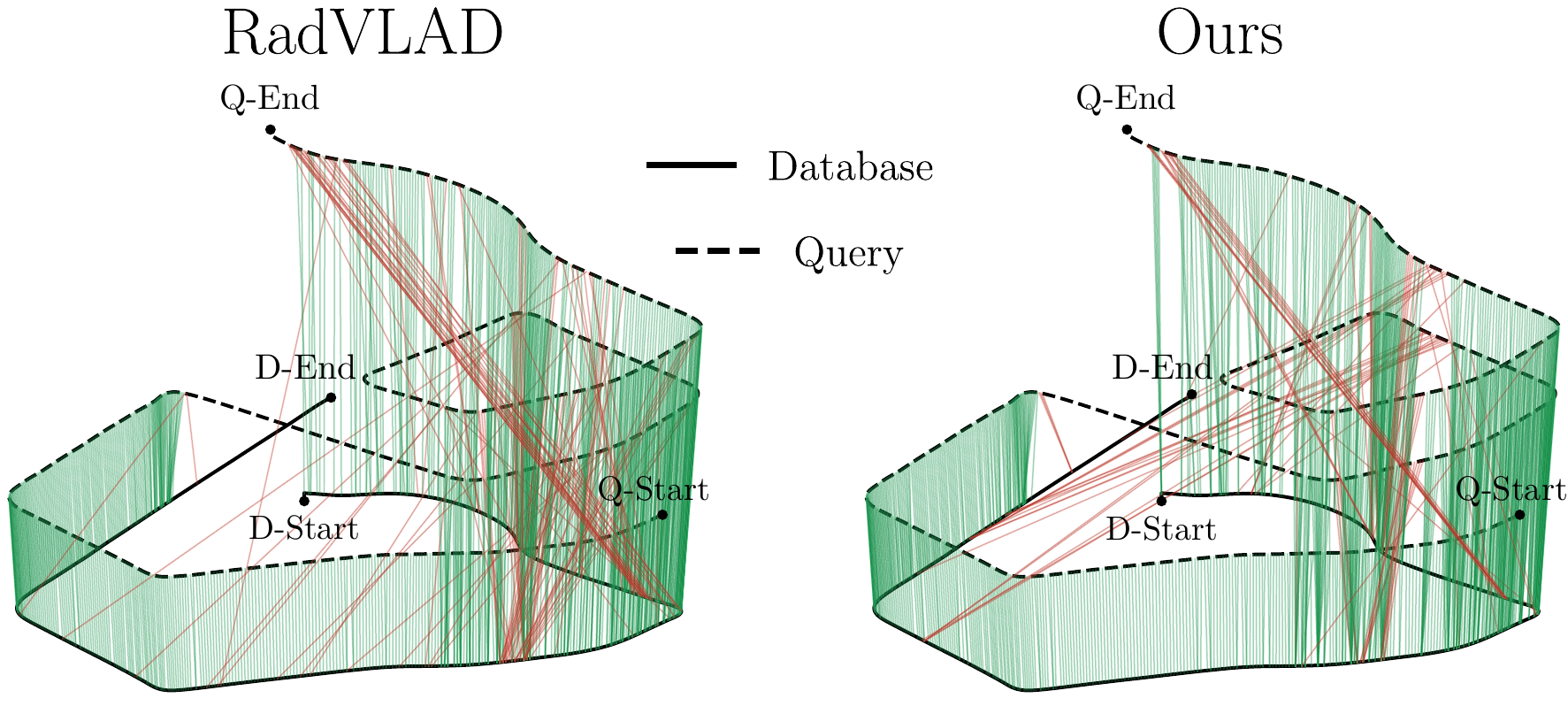}
    \caption{Qualitative comparison of trajectory-level retrieval consistency on Mulran DCC01~\cite{kim2020mulran}. {ReRadar uses Our (3); solid and dashed black curves denote database and query trajectories.} Retrieval links are shown for ReRadar and RadVLAD~\cite{gadd2024openradvlad}, where green indicates true positives and red indicates false positives.}
    \label{fig:placeholder}
\end{figure}

\begin{figure}[t!]
    \centering
    \includegraphics[width=0.48\textwidth]{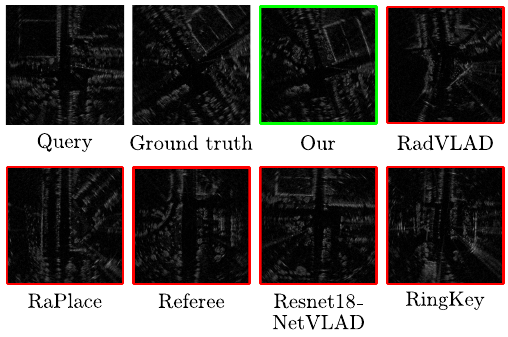}
    \caption{Qualitative comparison of retrieval results on a challenging query from Mulran KAIST03. {ReRadar results use the target-dataset-adapted model, Our (3).} The figure compares retrieval outputs from different methods for the same query.}
    \label{fig:compare}
\end{figure}

{We next evaluate adaptation across the two urban Mulran routes in Table~\ref{tab:mulran_result}. The benefit is most pronounced on DCC01: Our (1) matches RadVLAD in Recall@1, while cross-dataset adaptation in Our (2) increases Recall@1 and $F_1^{\max}$ to 81.28\% and 78.67\%, respectively. On KAIST03, both settings remain competitive, indicating that adaptation improves the difficult route without substantially degrading the easier one.

With target-dataset adaptation, Our (3) achieves the highest Recall@1 on both sequences, reaching 91.44\% on DCC01 and 98.33\% on KAIST03. Its $F_1^{\max}$ is also highest on DCC01, while remaining slightly below RadVLAD on KAIST03. Our (3) attains SR values of 88.24\% and 97.99\%; however, the lowest rotation and translation errors are distributed across different ReRadar settings, so adaptation does not improve every pose metric monotonically. Using the target-dataset-adapted model Our (3), we further visualize geometric registration, trajectory-level retrieval consistency, and a challenging correctly retrieved query in Figs.~\ref{fig:registration}--\ref{fig:compare}.}

\begin{table}[t!]
\centering

\caption{{Quantitative comparison on Mulran~\cite{kim2020mulran}. The ReRadar settings and evaluation metrics are defined in Sections~\ref{sec:setting} and~\ref{sec:metrics}, respectively. Best values are shown in bold.}}
\hypersetup{linkcolor=black,citecolor=black}
\resizebox{\linewidth}{!}{
\renewcommand{\arraystretch}{1.2}
\begin{tabular}{c|ccccc|ccccc}
 & \multicolumn{5}{c|}{DCC01}
 & \multicolumn{5}{c}{KAIST03} \\

Method
& Recall@1 & $F_1^{\max}$ & {SR} & $\hat{e_r}$ & $\hat{e_t}$
& Recall@1 & $F_1^{\max}$ & {SR} & $\hat{e_r}$ & $\hat{e_t}$ \\

\toprule
RaPlace~\cite{jang2023raplace}
& 64.71 & 66.47 & {--} & -- & -- & 92.98 & 94.70 & {--} & -- & -- \\

RadVLAD~\cite{gadd2024openradvlad}
& 75.94 & 76.55 & {--} & -- & -- & 96.66 & \textbf{97.95} & {--} & -- & -- \\

ResNet18-NetVLAD~\cite{gadd2024oord}
& 43.85 & 46.79 & {--} & -- & -- & 89.63 & 90.17 & {--} & -- & -- \\

ReFeree~\cite{kim2024referee}
& 54.55 & 56.46 & {44.39} & {1.74} & -- & 87.63 & 87.77 & {77.26} & {1.16} & -- \\

RingKey~\cite{kim2018scan}
& 54.01 & 54.88 & {--} & -- & -- & 86.29 & 86.53 & {--} & -- & -- \\

\midrule
{\textbf{Our (1)}}
& 75.94 & 74.30 & 75.40 & \textbf{1.46} & 5.40
& 95.32 & 93.12 & 94.98 & \textbf{0.79} & 0.85 \\

{\textbf{Our (2)}}
& 81.28 & {78.67} & 79.68 & 1.81 & 3.56
& 94.98 & {94.46} & 93.98 & 0.85 & 0.81 \\

{\textbf{Our (3)}}
& \textbf{91.44} & \textbf{80.99} & \textbf{88.24} & 1.66 & 3.87
& \textbf{98.33} & 97.47 & \textbf{97.99} & 0.81 & \textbf{0.79} \\

{\textbf{Our (2)-ResNet}}
& 75.94 &  72.73  & 70.05 & 1.97 & \textbf{2.49}
& 95.32 &  93.56 & 95.32 & 0.95 & 0.80 \\

{\textbf{Our (3)-ResNet}}
& 81.28 & 77.90 & 74.87 & 2.21 & 2.99
& 97.32 & 95.52 & 97.32 & 0.84 & 0.84 \\

\bottomrule
\end{tabular}
}

\label{tab:mulran_result}
\end{table}

 \subsection{\texorpdfstring{Results on Boreas}{Results on Boreas}}
\label{sec:boreas}

\begin{table}[t!]

\centering

\caption{{Quantitative comparison on Boreas~\cite{burnett2023boreas}. The ReRadar settings and evaluation metrics are defined in Sections~\ref{sec:setting} and~\ref{sec:metrics}, respectively. Best values are shown in bold.}}

\resizebox{\linewidth}{!}{
\renewcommand{\arraystretch}{1.2}
\begin{tabular}{c|ccccc|ccccc}
 & \multicolumn{5}{c|}{\texttt{2020-11-26-13-58}}
 & \multicolumn{5}{c}{\texttt{2021-01-26-11-22}} \\

Method
& Recall@1 & $F_1^{\max}$ & SR & $\hat{e_r}$ & $\hat{e_t}$
& Recall@1 & $F_1^{\max}$ & SR & $\hat{e_r}$ & $\hat{e_t}$ \\

\toprule
RaPlace~\cite{jang2023raplace}
& 94.93 & 94.93 & -- & -- & -- & 96.99 & 96.99 & -- & -- & -- \\

RadVLAD~\cite{gadd2024openradvlad}
& 91.30 & 91.80 & -- & -- & -- & 94.78 & 94.78 & -- & -- & -- \\

ResNet18-NetVLAD~\cite{gadd2024oord}
& 74.52 & 74.94 & -- & -- & -- & 73.98 & 73.98 & -- & -- & -- \\

ReFeree~\cite{kim2024referee}
& 73.67 & 73.90 & 65.38 & 1.26 & -- & 69.47 & 69.47 & 54.47 & 1.43 & -- \\

RingKey~\cite{kim2018scan}
& 58.33 & 58.33 & -- & -- & -- & 65.31 & 65.31 & -- & -- & -- \\

\midrule
\textbf{Our (1)}
& 86.85 & 86.85 & 86 & 0.72 & 3.64
& 77.54 & 78.49 & 77 & 0.80 & 3.77 \\

\textbf{Our (2)}
& 94.09 & 94.09 & 94.00 & 0.67 & 3.61
& 83.47 & 83.47 & 82.00 & \textbf{0.69} & 4.01 \\

\textbf{Our (3)}
& \textbf{97.35} & 97.35 & \textbf{97.00} & \textbf{0.59} & 3.47
& \textbf{99.38} & \textbf{99.43} & \textbf{99.00} & 0.75 & \textbf{3.39} \\

\textbf{Our (2)-ResNet}
& 87.70 & 87.70  & 87.00 & 0.75 & 3.73
& 83.29 & 83.37 & 83.00 & 0.82 & 3.67 \\

\textbf{Our (3)-ResNet}
& \textbf{97.35} & \textbf{97.40} & \textbf{97.00} & 0.66 & \textbf{3.37}
& 97.26 & 97.34 & 97.00 & 0.78 & 3.56 \\

\bottomrule
\end{tabular}
}

\label{tab:cross_dataset_boreas_result}

\end{table}
\hypersetup{linkcolor=black,citecolor=black}

{Finally, we assess long-term robustness under seasonal and weather changes using the two Boreas query sequences in Table~\ref{tab:cross_dataset_boreas_result}. The KITTI-only model Our (1) reaches 86.85\% and 77.54\% Recall@1. Adapting with Mulran data increases these values to 94.09\% and 83.47\%; thus, Our (2) nearly matches RaPlace on \texttt{2020-11-26-13-58}, but a larger gap remains on the more challenging \texttt{2021-01-26-11-22}.}

Target-dataset adaptation closes this gap. Our (3) reaches 97.35\% and 99.38\% Recall@1, surpassing the strongest external baseline on both sequences, and obtains SR values of 97\% and 99\%. Its mean rotation errors remain below one degree and its translation errors are below 3.5\,m, showing that the retrieved candidates also support accurate geometric pose estimation.

 \subsection{Computational Time}
\label{sec:time}

{For ReRadar on the OORD Bellmouth sequence with batch size one, the polar-to-Cartesian conversion, landmark-based pose-estimation, global-descriptor generation, and database-retrieval stages require 13.29, 190.00, 90.40, and 1.29\,ms on the Jetson Orin Nano (294.98\,ms total), and 1.45, 128.66, 18.85, and 0.75\,ms on the RTX 3090 (149.71\,ms total). Landmark-based pose estimation is the main bottleneck on both platforms.}
{\begin{figure}[t!]
    \centering
    \includegraphics[width=0.48\textwidth]{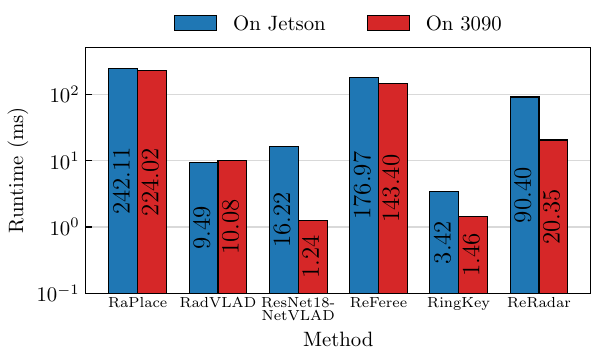}
    \caption{Global-descriptor generation runtime on Boreas \texttt{2021-03-02-13-38} using the Jetson Orin Nano and RTX 3090 (logarithmic scale).}
    \label{fig:runtime_comparison}
\end{figure}}

{ {For descriptor generation alone on Boreas \texttt{2021-03-02-13-38}, Fig.~\ref{fig:runtime_comparison} reports 90.40\,ms on the Jetson and 20.35\,ms on the RTX 3090. ReRadar is faster than RaPlace and ReFeree but slower than RadVLAD and RingKey. At the typical 4\,Hz radar rate, the end-to-end pipeline is close to real time, with further gains possible through optimization or parallelization.}}

{\subsection{Backbone Ablation}}
\label{sec:backbone_ablation}

{To isolate the effect of rotation-equivariant feature extraction, we compare Our (2) and Our (3) with their standard-ResNet counterparts under identical training settings. In the cross-dataset setting, ReResNet provides the clearest gains on difficult domain shifts: Recall@1 improves over Our (2)-ResNet by 53.05 percentage points on OORD Two Lochs, 5.34 points on Mulran DCC01, and 6.39 points on Boreas \texttt{2020-11-26-13-58}. On easier sequences such as Mulran KAIST03 and the second Boreas query, the two backbones are closer, indicating that equivariance is most beneficial when appearance and viewpoint changes are severe.}

{The same trend remains after target-dataset adaptation. Relative to Our (3)-ResNet, Our (3) improves Recall@1 by 4.51 points on OORD Two Lochs, 10.16 points on Mulran DCC01, and 2.12 points on Boreas \texttt{2021-01-26-11-22}. Standard ResNet can still match or slightly exceed ReResNet on some easier OORD pairs, and their pose errors are generally comparable. Thus, the principal benefit of ReResNet is stronger retrieval robustness on the most challenging cross-environment and long-term localization conditions rather than uniform improvement in every metric.}

{\subsection{Controlled Yaw-Rotation Evaluation}}
\label{sec:yaw_shift}

{To isolate the rotation-equivariant behavior of the intermediate feature maps, we conducted a controlled yaw-rotation evaluation using the ResNet and ReResNet variants trained on OORD Two Lochs 2. The trained models were kept fixed and evaluated separately on radar scans from OORD and Mulran.}

{For each scan $I_i$ and controlled yaw angle $\theta\in[0^\circ,360^\circ]$, we extracted local feature maps $F_i=f(I_i)$ and $F_{i,\theta}=f(\mathcal{R}_\theta I_i)$, where $\mathcal{R}_\theta$ denotes spatial rotation by $\theta$. To undo the applied yaw before comparison, the rotated feature map was inversely aligned as $\widetilde{F}_{i,\theta}=\mathcal{R}_{-\theta}\!\left(F_{i,\theta}\right)$. For each evaluation dataset $\mathcal{D}\in\{\mathrm{OORD},\mathrm{Mulran}\}$, the mean feature consistency was computed as}

{
\begin{equation}
\overline{S}_{\mathcal{D}}(\theta)
=\frac{1}{N_{\mathcal{D}}}\sum_{i=1}^{N_{\mathcal{D}}}
\frac{1}{|\Omega_{i,\theta}|}\sum_{p\in\Omega_{i,\theta}}
\frac{F_i(p)^\top\widetilde{F}_{i,\theta}(p)}
{\|F_i(p)\|_2\,\|\widetilde{F}_{i,\theta}(p)\|_2},
\end{equation}
}

{where $N_{\mathcal{D}}$ is the number of scans in dataset $\mathcal{D}$ and $\Omega_{i,\theta}$ excludes locations affected by rotation padding or cropping. A score near one indicates that inverse alignment recovers nearly the same local representation.}

\begin{figure}[t!]
    \centering
    \includegraphics[width=0.48\textwidth]{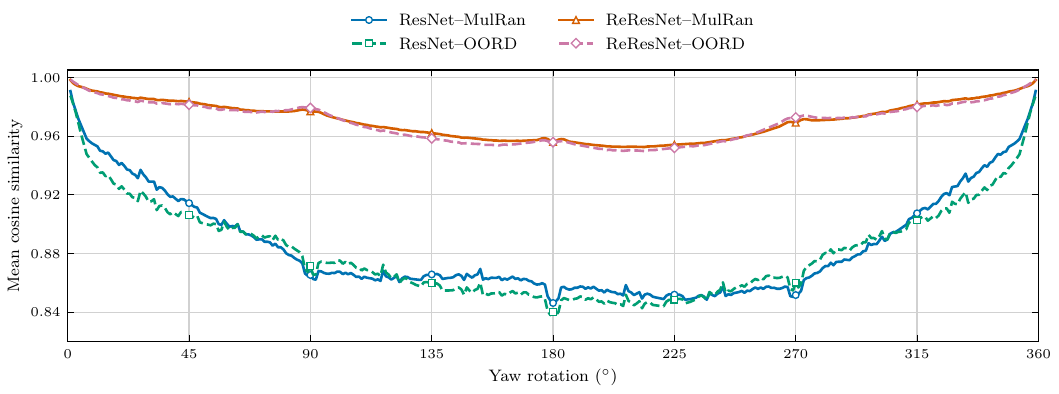}
    \caption{{Mean cosine consistency of spatially aligned local feature maps under controlled yaw rotations. ResNet and ReResNet are trained on OORD Two Lochs 2 and evaluated separately on OORD and Mulran.}}
    \label{fig:yaw_consistency}
\end{figure}

{We compare aligned local-feature consistency across the full yaw range in Fig.~\ref{fig:yaw_consistency}. On both datasets, ReResNet maintains cosine similarity above approximately 0.95, whereas ResNet decreases to approximately 0.84--0.87 for intermediate yaw rotations. The closely aligned OORD and Mulran curves indicate that the rotation-equivariant behavior learned from Two Lochs 2 transfers across datasets.}

\section{Conclusion}
\label{sec:conclusion}

In this paper, we presented a deep learning-based radar place-recognition and global-localization pipeline built on radar image representations. By repurposing radar-specific features, originally designed for ego-motion estimation, as keypoints for pose estimation and combining them with the InfoNCE objective for robust contrastive feature learning, the proposed method achieves strong performance on multiple public radar datasets.  {The comparison among KITTI-only, cross-dataset, and in-dataset training settings shows that radar-domain adaptation transfers effectively across datasets, while target-dataset adaptation provides further gains on the most challenging sequences. } {After target-dataset adaptation, ReRadar exceeds the strongest external baseline in Recall@1 on all evaluated OORD, Mulran, and Boreas database--query pairs.} Despite these encouraging results, there is further space for improvement. Our current pipeline still has notable runtime overhead in the {landmark-based pose-estimation stage}. In future work, we will improve inference speed by reducing the runtime of this stage and designing more lightweight descriptors.

\bibliographystyle{ieeetr}
\bibliography{references}

\end{document}